# A STUDY FOR THE EFFECT OF THE EMPHATICNESS AND LANGUAGE AND DIALECT FOR VOICE ONSET TIME (VOT) IN MODERN STANDARD ARABIC (MSA)

Sulaiman S. AlDahri

King Abdulaziz City for Science and Technology, Riyadh, Saudi Arabia

saldahri@kacst.edu.sa

## ABSTRACT

*The signal sound contains many different features, including Voice Onset Time (VOT), which is a very important feature of stop sounds in many languages. The only application of VOT values is stopping phoneme subsets. This subset of consonant sounds is stop phonemes exist in the Arabic language, and in fact, all languages. Very important subsets of Semitic language's consonants are the Emphatic sounds. The pronunciation of these sounds is hard and unique especially for less-educated Arabs and non-native Arabic speakers. In the Arabic language, all emphatic sounds have their own non-emphatic counterparts that differ only in the "emphaticness" based on written letters. VOT can be utilized by the human auditory system to distinguish between voiced and unvoiced stops such as /p/ and /b/ in English. Similarly, VOT can be adopted by digital systems to classify and recognize stop sounds and their carried syllables for words of any language. In addition, an analysis of any language's phoneme set is very important in order to identify the features of digital speech and language for automatic recognition, synthesis, processing, and communication.*

*The main reason to choose this subject is that there is not enough research that analyzes the Arabic language. Also, this subject is new because it will analyze Modern Standard Arabic (MSA) and other Arabic dialects.*

*This search focuses on computing and analyzing VOT of Modern Standard Arabic (MSA), within the Arabic language, for all pairs of non-emphatic (namely, /d/ and /t/) and emphatic pairs (namely, /d$^?$/ and /t$^?$/) depending on carrier words. This research uses a database built by ourselves, and uses the carrier words syllable structure: CV-CV-CV.*

*One of the main outcomes always found is the emphatic sounds (/d$^?$/, /t$^?$/) are less than 50% of non-emphatic (counter-part) sounds ( /d/, /t/).Also, VOT can be used to classify or detect for a dialect ina language.*

## 1. INTRODUCTION

### 1.1. Voice Onset Time

The VOT is a main feature used to differ between voice stops and unvoiced stops. Phonation onset or VOT is defined as the length time (period) between the onset of voicing pulses and the release of the primary occlusion of the vocal tract as can be seen in Figure 1.1. VOT as we have just described is relevant only for stop consonants [1] [2]. This period is usually measured in milliseconds. Stop consonants are produced with a closure of the vocal tract at a specific place which is known as the place of articulation [3].

There are three types of VOT. There are zero VOT, positive VOT and negative VOT. Zero VOT means where the onset of vocal fold vibration coincides (approximately) with the plosive release. Positive VOT means that there is a delay in the onset of vocal fold vibration after the plosive release. Negative VOT the onset of vocal fold vibration precedes the plosive release [1] [4].

VOT is an important characteristic of stop consonants. It plays a great role in perceptual discrimination of phonemes of the same place of articulation [5]. It is also used in word segmentation, stress related phenomena, and dialectal and accented variations in speech patterns [1].Moreover, previous research found VOT values are not affected by the change of gender of the addressee [6][7].

In languages which process two categories of voicing, there are two types of sounds: voiced and voiceless. Depend on VOT, Liskeret al. [1] divided languages into two groups: group A languages which have long VOT, over 50 milliseconds, for a voiceless stop but short VOT for voiced: and group B languages which have short VOT, less than 30 milliseconds, for voiceless, but negative VOT for voiced [1][3].

Reports on this topic in Arabic are not uniform. According to Al-Ani's data [8] and Mitleb's data [9] Arabic is a member of Group A while Yeni-Komshian et al. [10], show that Arabic belongs to Group B. Flege [11] considers that Arabic neither belongs to Group A nor Group B.VOT values are generally unobserved in fixed-length frame-based speech investigation. On the other hand, it is known that VOT can help enhance the performance of automatic speech recognition (ASR). Among the various applications of the use of VOT is the difficulty of accent detection. Non-native language can affect both the length and the quality of the VOT of English stops [12]. Depending on a research effort [12], VOT values can be used to discriminate Mandarin, Turkish, German, Spanish and English accents.

## 1.2. Literature review

### 1.2.1. VOT across languages

The VOT of languages spoken in industrial countries, mainly English, Japanese, and German, have been researched for more than forty years. Lisker et al. [1] investigated VOT for more than nine languages and dialects under different environments. Among these languages were English that was studied using American and Britain dialects. Lisker et al. found that the perceptual relevance of the timing adjustments of the glottal gap are significant to the articulation for phonological distinctions in different languages. Also, Lisker found that the listener best discriminates variants along these temporal dimension sat the boundary zones between sound categories. There have been several studies in English which show results similar to those of Lisker et al [1]. Peterson et al. [13] present their VOT results of /p/, /t/, /k/ as 58 milliseconds, 69 milliseconds and 75 milliseconds respectively. Flege et al. [14] found the VOT of /p/ is 46 milliseconds, /t/ is 56 milliseconds and /k/ is 67 milliseconds [3].

In another effort, Das et al. [12] tried to detect VOT values for unvoiced stops (/p, t, k/) using the Teager energy operator for automatic detection of accented English. They mainly applied their algorithm to accent classification using English, Chinese, and Indian accented speakers. Among the 546 tokens consisting of 3 words from 12 speakers, their average mismatch between automatic and hand labeled VOT was 0.735 milliseconds. This represented a 1.15% mismatch. Also, they proved that the average VOT values are different among three different language groups, hence making VOT values a good feature for accent classification.

To be more specific about the English language, the VOT values for /d/ are ranging between 0 milliseconds to -155 milliseconds. For /t/ values are ranging between 30 milliseconds to 105 milliseconds. For /b/ values are ranging between 0 milliseconds to -130 milliseconds. For /p/ values are ranging between 20 milliseconds to 120 milliseconds. For /g/ values are ranging between 0 milliseconds to -150 milliseconds. For /k/ values are ranging between 50 milliseconds to 135 milliseconds [1].

### 1.2.2. Arabic language VOT research

There is a glaring lack of modern research on the Arabic language in the fields of references and resources regarding digital speech and language processing [3]. One of the important areas of researches in any language is the investigation of the VOT values of its stops. A few researchers have investigated VOT in Arabic. The first research was conducted by Alghamdi [2] and analyzed the role of VOT in speaker identification and the effect of acquiring a second language on the Ghamdi analysis of Saudi speaker stops' vocalizations. His research showed the presence of individual differences among Arabic speakers in terms of VOT. Also, he showed that a phonetic diversity between the first language and the second language is maximized when the speakers are more fluent in the second language. In other words, he emphasized that it can be predicted from Arabic speech that the speaker is fluent in a foreign language with long VOT values. Moreover, Alghamdi [2] investigated that for a Saudi dialect in the Arabic language, the results of average VOT for /t/, a /k/ and /t?/ are 39 milliseconds, 42 milliseconds and 21 milliseconds, respectively.

In another study, Mitleb [9] analyzed VOT of Jordanian Arabic stops. One of his results is that the VOT value is dependent on vowel length, where with long vowel environment the VOT is harder compared to short vowel environment. Also, he realized that VOT distinguishes Arabic's unvoiced and voiced stops as is the case in English. Also, he found that the Arabic unvoiced alveolar stop /t/ is not different from the unvoiced velar stop /k/ with regard to VOT values.

Mitleb's [9] findings about an Arabic Jordanian accent VOT values are as follows: for neighbouring short vowel /I/, /d/, VOT value is 10 milliseconds; for /t/, 37 milliseconds; for /k/, 39 milliseconds; and for /g/, 15 milliseconds. In addition to this, in case of neighbouring long vowel /I: /, the long vowel for /d/ is 23 milliseconds, for/t/, 64 milliseconds, for /k/, 60 milliseconds and for /g/, 20 milliseconds. Also, in Alghamdi's [3] experiment, he found for the Ghamidi dialect of the Arabic language, the results of average VOT for /t/ and /k/ in the initial position of the word, the follow the two /a/ vowels are 25 milliseconds and 30.3 milliseconds. In AlDahri's [15] experiment, he found that the /d/ VOT values in Modern Standard Arabic (MSA) range between 12 and 22 milliseconds and the /t/ VOT values in MSA range between 38 and 93 milliseconds. In addition, he concluded the VOT values of these stops (/t/,/d/) are positive regardless of the voicing, where /d/ is voiced sound, but /t/ is not. This is not the case for the same sounds in the English language where voiced stops have negative VOT values, but unvoiced (e.g., /t/) have positive VOT values. [16]. Another, in AlDahri's [17] experiment, he investigated the four MSA Arabic stops namely /d/, /d$^?$/, /t/ and /t?/ by analyzing their VOT values. We ended to a conclusion of that VOT values of these stops are positive. In addition, he realized the fact that VOT values of /d/ and /t/ phonemes are always more than VOT values of /d?/ and /t?/ phonemes. Also, we found the standard deviation for non-emphatic phonemes is higher than that of their emphatic counterparts by about three times. This implies the high variability and difficulty of pronunciation for emphatic phonemes. Finally, in AlDahri's experiment, [18] he investigated two main standards in Arabic language which are MSA Arabic and CA Arabic by computing, analyzing and comparing the VOT. He found that for the MSA and CA Arabic, voiced sounds have short VOT while the unvoiced sounds have long VOT. In addition, he found that VOT values vary from one Arabic dialect to another. This shows that VOT can be used for dialect classification or detection.

### 1.3. Arabic language Overview

Arabic is a Semitic language, and it is one of the oldest languages in the world. Currently, it is the second language in terms of the number of speakers [18]. Arabic is the first language in the Arab world, i.e., Saudi Arabia, Jordan, Yemen, Egypt, Syria, Lebanon, etc. Arabic alphabets are used in several languages, such as Persian and Urdu. The MSA consists of 34 sounds: 28 consonants and 6 vowels [19]. It has three long vowels (/i: /, /a: /, /u: /) and three short vowels (/i/, /a/, /u/), while American English has twelve vowels [20]. The Arabic language has fewer vowels than the English language.

However, the sound is the smallest element of a speech unit that indicates a difference in meaning, word, or sentence. Arabic sounds contain two distinct classes. They are pharyngeal and emphatic sounds. These two classes can be found only in a Semitic language like Hebrew [19][21]. The allowed syllables in the Arabic language are: CV, CVC, and CVCC where V indicates a (long or short) vowel while C indicates a consonant. Arabic utterances can only start with a consonant [19]. All Arabic syllables must contain at least one vowel. Also Arabic vowels cannot be initialled and they can occur either between two consonants or be the final sound of a word. Arabic syllables can be classified as short or long. All vowels that exist in MSA also exist in Classical Arabic (CA). The CV type is short while all others are long. Syllables can also be classified as open or closed. An open syllable ends with a vowel while a closed syllable ends with a consonant [22].

MSA is widely taught in schools, universities, and used in workplaces, government and the media. MSA derives from CA, the only surviving member of the Old North Arabian dialect group, found in Pre-Islamic Arabic inscriptions dating back to the 4th century. CA has also been a literary language and the liturgical language of Islam since its inception in the $^7$th century. MSA, Standard Arabic, or Literary Arabic is the standard and literary variety of Arabic used in writing and in formal speech. Most western scholars distinguish two standard varieties of the Arabic language: the CA of the Qur'an and early Islamic ($^7$th to $^9$th centuries) literature, and MSA which is the standard language used today [23].

### 1.4. Emphatic Consonants in MSA Arabic

In the case of the Semitic languages, the emphasis is a phonetic feature characterizing a consonant. There are four emphatic phonemes in MSA Arabic as can be seen in Table 1. Also, some researchers [19][23] added /lʕ/ phoneme in word /ʔalla:h/ to emphatic phonemes. An interesting fact about Arabic is that it is the only language that contains the emphatic phoneme "dhaad" /dʕ/ and hence Arabic is also alternatively called the "dhaad language" because of this uniqueness [24]. An emphatic phoneme that is very similar to /dʕ/ is /ðʕ/. Some people nowadays, including some native speakers, have some confusion in uttering and recognizing these two phonemes. This factor adds more complexity to machine-based recognition, synthesis, and manipulation of the Arabic language because if humans face difficulties in dealing with these phonemes, it will imply more and more machine shortcomings and lack of knowledge [25].

Table 1. The Emphatic and non-Emphatic counterpart sounds in MSA Arabic language

| Arabic Alphabet Carrier | IPA Symbol | Non-Emphatic Counterpart |
|---|---|---|
| Daad ض | $d^ʕ$ | /d/ Daal |
| Saad ص | $s^ʕ$ | Voice: /z/ (zain); Unvoiced: /s/ (Seen) |
| T_aa ط | $t^ʕ$ | Voice: /d/ (Daal); Unvoiced: /t/ (Taa) |
| Dhaa ظ | $ð^ʕ$ | /ð/ (Thaal) |

This journal presents the research work of a student earning his masters degree. His study analyse the VOT for four MSA Arabic stops sounds /d/, /t/, /dʕ/ and /tʕ/. The rest of the journal is organized as follows: Section 2 will present a description about the used corpus and the experimental set up. Section 3 will give the results of the research in addition to some discussions. Before the final section, Section 4 will summarize the conclusions of the research. Finally, the Section 5 will list our references.

## 2. EXPERIMENTAL FRAMEWORK

The set of stop sounds in the Arabic Language consists of eight sounds and we can classify them into: emphatic and non-emphatic or voiced and unvoiced [26]. These sets are best illustrated in Table 2 with a full description of their place of articulation, voicing, and emphasis properties.

Table 2. Stop sounds in Arabic language

|      |          |              | Bilabial | Alveo-dental | Velar | Uvular | Glottal |
|------|----------|--------------|----------|--------------|-------|--------|---------|
| Stop | Voiced   | Emphatic     |          | /dˤ/         |       |        |         |
|      |          | Non-emphatic | /b/      | /d/          |       |        |         |
|      | Unvoiced | Emphatic     |          | /tˤ/         |       |        |         |
|      |          | Non-emphatic |          | /t/          | /k/   | /q/    | /ʔ/     |

### 2.1. Used Speech Corpus

As a fundamental step for this work, our search depends on a corpus of words built with the seven targeted phonemes. These carrier words which were used in our search are clear as we see in Table 3. This corpus took seven weeks to be built. Delivering a high quality corpus will save the time and effort of the researchers who are going to conduct similar work. Since the corpus is very critical to ensure the quality of the result. High attention is paid to ensure the quality of speakers' pronunciation skills and recording clarity.

The speakers who participate in this corpus are selected carefully in order to satisfy the utterance quality required for this work. The best Arabic speakers who can pronounce unique MSA sounds correctly are those who master The Holy Quran (THQ). Therefore, the speakers we selected for this corpus should not have confusion in pronunciation.

Table 3. Investigated sounds with carrier words information in the corpus

| Arabic Alphabet Carrier | IPA Symbol | Carrier words (CV-CV-CV) | Transcription | Code |
|---|---|---|---|---|
| Daal    د | /d/  | ندر | /nadara/   | C1e0 |
| Dhaad   ض | /dˤ/ | نضر | /nadˤara/  | C1e1 |
| Taa     ت | /t/  | نتر | /natara/   | C3e0 |
| T_aa    ط | /tˤ/ | نطر | /natˤara/  | C3e1 |
| Kaaf    ك | /K/  | نكر | /nakara/   | C5e0 |
| Qaaf    ق | /q/  | نقر | /naqara/   | C6e0 |
| Baa     ب | /b/  | نبر | /nabara/   | C7e0 |

Among the Arabic speakers, people who master recitation of THQ are guaranteed to not have this confusion. Thus sixty male and female speakers are selected and they master the recitation of THQ. They are native and non-native Arabic speakers. The ages of these speakers range between thirteen and forty years old.

When recording the corpus, each speaker utters seven words carrying the phonemes to be analyzed as we see in Table 3. These words are chosen to make sure that the targeted phonemes are in the middle of the word while the preceding and the succeeding phonemes with respect to the targeted phonemes are always the same. It is the short vowel /a/. The words structure is CV-CV-CV. The speaker repeats this set of words for five trials. Therefore, the total number of the recorded utterances is 2100 (60 speakers × 7 words × 5 trials = 2100 recorded words).For the recording we set the sampling rate at 16000 sample/second(16 kHz) and resolution at 16 bit using one channel (mono).

## 2.2. Files Coding

In order to organize the research and ease tracking, managing our results and conclusions, the audio file names have been coded in specific formats. Each audio file is named according to the following naming pattern: SxxCyEzTw.wav. In this string S, C, E and T stand for speaker, consonant, emphatic, and trial, respectively. The 'xx' (two digits number) displays the speaker number. The one digit 'y' is the emphatic/non-emphatic sound identifier as follows: 1 refers to the pair /d$^?$/ or /d/ , 3 refers to the pair /t$^?$/ and /t/, 5 refers to the pair /k/, 6 refers to the pair /q/, and finally 7 refers to the pair /b/. The fourth digit 'z' is a binary flag set to 0 for non-emphatic and 1 for emphatic. The last digit 'w' is a one-digit number representing the trial number. These sets are best illustrated in Table 3.

## 2.3. Methodology

To achieve our objectives, the research of our experiment depends mainly on extracting VOT values of the Emphatic and non-Emphatic phonemes in MSA Arabic stops. Our analysis was implemented by using Wavesurfer tools [27] and spectrograms' readings.

We used signal energy and vocal cord vibration information (i.e., fundamental frequency) to locate the beginning of stop release, closure, and voicing. In both cases for Emphatic and non-Emphatic stop phonemes, we found positive VOT in values in MSA Arabic, a stop which means that voicing occurs only after the closure release.

The closure release was measured from the beginning of the abrupt increase in the energy level as can be read from the Wavesurfer signal analysis and our own designed spectrograms. Voicing onset (i.e., start of vocal cords vibration) can be observed first by noticing low frequency periodicity in the Wide-band spectrograms which can be seen as vertical lines.

# 3. RESULTS

In this Section, we will investigate our goals: Comparing and analyzing voiced/unvoiced sounds which are /d/ and /t/, Comparing and analyzing Emphatic sounds which are /t$^?$/ and /d$^?$/, Studying the gender effect, Studying the Memorization effect, The effect of emphasis in MSA Arabic language, The effect of VOT in different MSA Arabic dialects.

## 3.1. Comparing and analyzing voiced/unvoiced sounds /d/ and /t/

In this section, we complete the previous research [15]. The initial outcomes from our investigation regarding MSA Arabic /d/ and /t/ sounds are listed in Table 4. We investigated many audio files in our corpus, but the table listed VOT values of twenty audio files for /d/ sound and another twenty audio files for /t/ sound. The /d/ VOT values are ranging between 14 and 22 milliseconds and /t/ VOT is ranging between 32 and 71 milliseconds. Also, as we can see from Figure 1, the averages of the VOT values for /d/ and /t/ are 16 and 51.65 milliseconds, respectively. Averages, standard deviations, maximum, and minimum of the VOT values for these sounds can certify that this feature can be used to distinguish these two sounds in MSA Arabic.

Table 4. Computed VOT values of /d/ and /t/ sounds

| /d/ | | /t/ | |
|---|---|---|---|
| **Audio File** | **VOT (mSec)** | **Audio File** | **VOT (mSec)** |
| s01c1e0t1 | 17 | s01c3e0t1 | 51 |
| s01c1e0t2 | 19 | s01c3e0t2 | 57 |
| s02c1e0t1 | 22 | s02c3e0t1 | 61 |
| s02c1e0t2 | 17 | s02c3e0t2 | 48 |
| s04c1e0t1 | 16 | s04c3e0t1 | 59 |
| s04c1e0t2 | 14 | s04c3e0t2 | 57 |
| s06c1e0t1 | 16 | s06c3e0t1 | 33 |
| s06c1e0t2 | 14 | s06c3e0t2 | 36 |
| s07c1e0t2 | 18 | s07c3e0t2 | 54 |
| s07c1e0t3 | 16 | s07c3e0t3 | 59 |
| s09c1e0t1 | 14 | s09c3e0t2 | 55 |
| s09c1e0t2 | 14 | s09c3e0t3 | 50 |
| s014c1e0t1 | 14 | s14c3e0t1 | 71 |
| s014c1e0t2 | 15 | s14c3e0t2 | 69 |
| s017c1e0t2 | 14 | s17c3e0t2 | 34 |
| s017c1e0t3 | 14 | s17c3e0t3 | 32 |
| s019c1e0t2 | 16 | s19c3e0t2 | 50 |
| s019c1e0t3 | 15 | s19c3e0t3 | 50 |
| s021c1e0t1 | 16 | s21c3e0t1 | 55 |
| s021c1e0t2 | 19 | s21c3e0t2 | 52 |

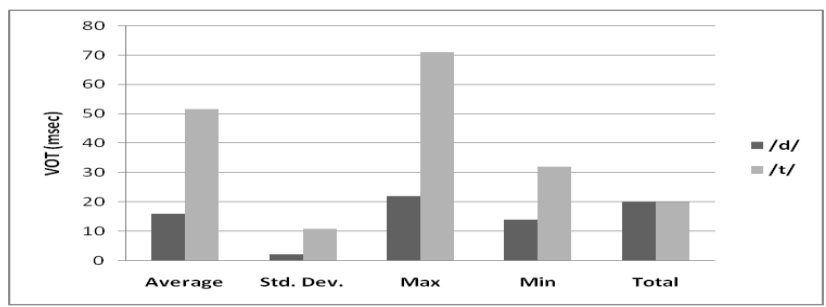

Figure 1. Some derived statistics about VOT values

In addition, we noticed that the VOT values for both the voiced and unvoiced Arabic MSA stops are positive. This is not the case in other languages such as English where voiced stops have negative VOT values as reported in [1]. English voiced stops have negative VOT values whereas unvoiced ones have positive VOT values as reported by [1]. Negative VOT values imply that vocal cords start vibrating before the vocal tract releases while uttering stops. This is contrary to the MSA Arabic stops.

We also noticed that VOT values for Arabic stops are dependent on the dialect and also dependent on acquiring the second language as confirmation to outcomes reported by Alghamdi [2]. One of our conclusions here is that the VOT of Arabic stops can be used to distinguish between the Arabic language and other languages because it is easy to locate stops in any speech segment. In addition, VOT values can be used to recognize the dialect of the speaker.

Table 5 shows the VOT values of /d/ and /t/ stops for the Arabic language with three dialects as well as their values for English as reported by four references [1][2][9][14]. It can be noticed that all Arabic dialects in the table have no negative VOT values for voiced stops, unlike English, depending on more than one researcher [1][2][12][14]. Among the values of the three Arabic dialects presented in the table, we can notice less variation of VOT values of /d/ sounds which is a voiced stop.

Table 5. The average of VOT values of /d/ and /t/ for some dialects and languages

| Reference | /d/ VOT (msec) | /t/ VOT (msec) | Comments |
|---|---|---|---|
| **Mitleb (short vowel)** | 10 | 37 | jordanian dialect |
| **Mitleb (long vowel)** | 24 | 64 | jordanian dialect |
| **Algamdi** | ** | 30 | Saudi dialect |
| **Lisker and Abramson (1964)** | -102 | 70 | English Language |
| **Our study** | 16 | 51.65 | MSA Arabic (Average) |

On the other hand, there are wide variations of /t/ VOT values among different Arabic dialects. This means that Arabic speakers mostly vocalize /d/ stops in all three dialects in the same way, at least regarding VOT values. On the contrary, Arabic speakers of these three dialects have big variations of vocalizing the /t/ stop which is an unvoiced stop.

Regarding the /t/ stop, we can conclude that there is a wide variation in articulating and hearing in the English language in both of its main dialects, American English and British English. In the same way, Arabic dialects have noticeable distinctions from both the perception and vocalization points of views.

In conclusion, the VOT values of these stops are always positive, regardless of the voicing where /d/ is a voiced sound, but /t/ is not. Using the work of previous researchers, we compared VOT values to values in other Arabic dialects. We noticed that the VOT value of the /t/ sound was more dependent of the different Arabic dialects, while the /d/ has less dependency.

## 3.2. Analyzing Emphatic sounds /dˤ/ and /tˤ/

The initial outcomes from our investigation regarding MSA Arabic /d/ and /t/ sounds are listed in Table 6. We investigated many audio files in our corpus, but the table listed VOT values of twenty audio files for /dʔ/ sound and another twenty audio files for /tˤ/ sound. The /dˤ/ VOT values are ranging between nine and fourteen milliseconds and /tˤ/ VOT is ranging between 15 and 24 milliseconds. Also, we can see from Figure 2, the averages of the VOT values for /dˤ/ and /tˤ/ are 11.5 and 18.35 milliseconds, respectively. Averages, standard deviations, maximum, and minimum of the VOT values for these sounds can certify that this feature can be used to distinguish these two sounds in MSA Arabic.

Table 6. Computed VOT values of /dˤ/ and /tˤ/ sounds

| /d?/ | | /t?/ | |
|---|---|---|---|
| Audio File | VOT (mSec) | Audio File | VOT (mSec) |
| s01c1e1t1 | 10 | s01c3e1t2 | 16 |
| s01c1e1t2 | 9 | s01c3e1t3 | 15 |
| s02c1e1t1 | 10 | s02c3e1t1 | 23 |
| s02c1e1t2 | 12 | s02c3e1t2 | 24 |
| s04c1e1t1 | 14 | s04c3e1t1 | 22 |
| s04c1e1t2 | 14 | s04c3e1t2 | 24 |
| s06c1e1t2 | 10 | s06c3e1t2 | 18 |
| s06c1e1t3 | 12 | s06c3e1t3 | 18 |
| s07c1e1t2 | 9 | s07c3e1t2 | 17 |
| s07c1e1t3 | 9 | s07c3e1t3 | 17 |
| s09c1e1t2 | 13 | s09c3e1t2 | 16 |
| s09c1e1t3 | 12 | s09c3e1t3 | 15 |
| s14c1e1t1 | 11 | s14c3e1t1 | 17 |
| s14c1e1t2 | 10 | s14c3e1t2 | 15 |
| s17c1e1t2 | 13 | s17c3e1t2 | 20 |
| s17c1e1t3 | 12 | s17c3e1t3 | 19 |
| s19c1e1t2 | 12 | s19c3e1t2 | 15 |
| s19c1e1t3 | 14 | s19c3e1t3 | 17 |
| s21c1e1t1 | 12 | s21c3e1t1 | 19 |
| s21c1e1t2 | 12 | s21c3e1t2 | 20 |

In another words, when we compared the averages of VOT values for emphatic sounds, we always found the VOT for voiced sound /tˤ/ is more than unvoiced sound /dˤ/. Also, we found there was no overlapping in the average VOT. Both were 7 milliseconds.

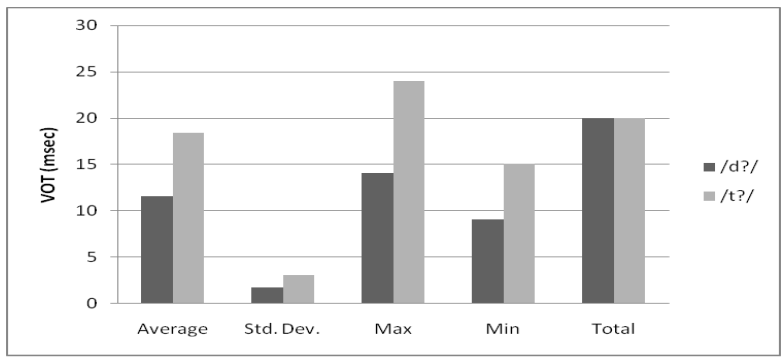

Figure 2. Some derived statistics about VOT values

The conclusion that we came up with is that the VOT values of these stops are positive regardless of the voicing where /t$^?$/ is voiced sound, but /d$^?$/ is not. Also, we found that the VOT for voiced sound /t$^?$/ is more than unvoiced sound /d$^?$/.

### 3.3. The effect of Emphasis in MSA Arabic language

In this section, we complete the previous research [17]. The initial outcomes from our investigation regarding MSA Arabic /d/ and /d$^?$/ sounds are listed in Table 8 and /t/ and /t$^?$/ sounds are listed in Table 9. We investigated many audio files in our corpus, but the table listed VOT values of 20 audio files for /d/ sound and another twenty audio files for /d?/ sound and the same thing /t/ sound and /t$^?$/ sound. The /d/ VOT values are ranging between 14 and 22 milliseconds and /d$^?$/ VOT is ranging between 9 and 14 milliseconds. On the other side, the /t/ VOT values are ranging between 32 and 71 milliseconds and /t$^?$/ VOT is ranging between 13 and 25 milliseconds.

Table 8. Computed VOT values of /d/ and /d$^?$/ sounds

| /d/ | | /d?/ | |
| --- | --- | --- | --- |
| **Audio File** | **VOT (mSec)** | **Audio File** | **VOT (mSec)** |
| s01c1e0t1 | 17 | s01c1e1t1 | 10 |
| s01c1e0t2 | 19 | s01c1e1t2 | 9 |
| s02c1e0t1 | 22 | s02c1e1t1 | 10 |
| s02c1e0t2 | 17 | s02c1e1t2 | 12 |
| s04c1e0t1 | 16 | s04c1e1t1 | 14 |
| s04c1e0t2 | 14 | s04c1e1t2 | 14 |
| s06c1e0t1 | 16 | s06c1e1t2 | 10 |
| s06c1e0t2 | 14 | s06c1e1t3 | 12 |
| s07c1e0t2 | 18 | s07c1e1t2 | 9 |
| s07c1e0t3 | 16 | s07c1e1t3 | 9 |
| s09c1e0t1 | 14 | s09c1e1t2 | 13 |
| s09c1e0t2 | 14 | s09c1e1t3 | 12 |
| s014c1e0t1 | 14 | s14c1e1t1 | 11 |
| s014c1e0t2 | 15 | s14c1e1t2 | 10 |
| s017c1e0t2 | 14 | s17c1e1t2 | 13 |
| s017c1e0t3 | 14 | s17c1e1t3 | 12 |
| s019c1e0t2 | 16 | s19c1e1t2 | 12 |
| s019c1e0t3 | 15 | s19c1e1t3 | 14 |
| s021c1e0t1 | 16 | s21c1e1t1 | 12 |
| s021c1e0t2 | 19 | s21c1e1t2 | 12 |

Table 9. Computed VOT values of /t/ and /t$^?$/ sounds

| /t/ | | /t?/ | |
|---|---|---|---|
| **Audio File** | **VOT (mSec)** | **Audio File** | **VOT (mSec)** |
| s01c3e0t1 | 51 | s01c3e1t2 | 16 |
| s01c3e0t2 | 57 | s01c3e1t3 | 15 |
| s02c3e0t1 | 61 | s02c3e1t1 | 23 |
| s02c3e0t2 | 48 | s02c3e1t2 | 24 |
| s04c3e0t1 | 59 | s04c3e1t1 | 22 |
| s04c3e0t2 | 57 | s04c3e1t2 | 24 |
| s06c3e0t1 | 33 | s06c3e1t2 | 18 |
| s06c3e0t2 | 36 | s06c3e1t3 | 18 |
| s07c3e0t2 | 54 | s07c3e1t2 | 17 |
| s07c3e0t3 | 59 | s07c3e1t3 | 17 |
| s09c3e0t2 | 55 | s09c3e1t2 | 16 |
| s09c3e0t3 | 50 | s09c3e1t3 | 15 |
| s14c3e0t1 | 71 | s14c3e1t1 | 17 |
| s14c3e0t2 | 69 | s14c3e1t2 | 15 |
| s17c3e0t2 | 34 | s17c3e1t2 | 20 |
| s17c3e0t3 | 32 | s17c3e1t3 | 19 |
| s19c3e0t2 | 50 | s19c3e1t2 | 15 |
| s19c3e0t3 | 50 | s19c3e1t3 | 17 |
| s21c3e0t1 | 55 | s21c3e1t1 | 19 |
| s21c3e0t2 | 52 | s21c3e1t2 | 20 |

In addition, we can see from Figure 4, the averages of the VOT values for /d/ and /d$^?$/ are 16 and 11.5 milliseconds, respectively. Averages, standard deviations, maximum, and minimum of the VOT values for these sounds can certify that this feature can be used to distinguish these two sounds in MSA Arabic. Also, from Figure 5, we can use it to distinguish between the two sounds /t/ and /t$^?$/ in MSA Arabic.

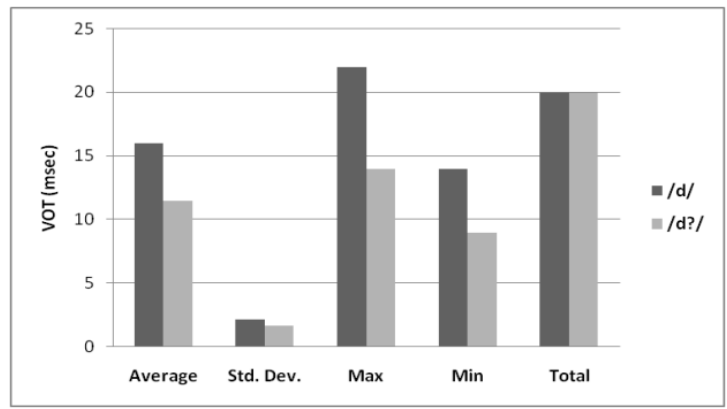

Figure 4. Some derived statistics about VOT values

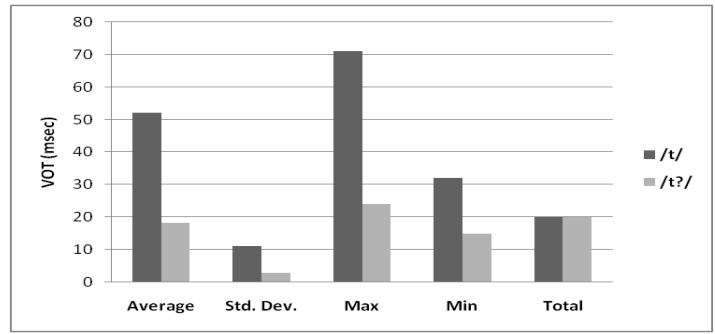

Figure 5. Some derived statistics about VOT values

Finally to conclude, we found that the VOT value for non emphatic stop sounds /d/ and /t/ are always more than VOT values for emphatic stop sounds /d$^?$/ and /t$^?$/. This is compatible with the previous unpublished PhD thesis [29]. In other words, the difference between the average VOT value for /d/ and /d$^?$/ sounds is 4 milliseconds. It means the value VOT for /d$^?$/ is almost equal to 70% of VOT value of /d/. Also, the difference between the average VOT value for /t/ and /t$^?$/ sounds is 34 milliseconds. It means the value VOT for /t/ is almost equal to 200% of the VOT value of /t$^?$/.

In addition, we noticed that the VOT values for both the emphatic and non-emphatic Arabic MSA stops are positive. In other words, we noticed that the voiced sounds (/d/ and /d$^?$/) and unvoiced sounds (/t/ and /t$^?$/) always have positive VOT values. This is not the case in other languages such as English and Spanish where voiced stops have negative VOT values as reported in [1]. Spanish voiced stops have negative VOT values whereas unvoiced also have negative VOT values as reported by [1]. Negative VOT values imply that vocal cords start vibrating before the vocal tract releases, while uttering stops. This is contrary to MSA Arabic stops.

One of our conclusions is that the VOT of Arabic stops can be used to distinguish between emphatic sounds and non-emphatic sounds.

Moreover, we found the standard division for non-emphatic sounds is more than the standard deviation for emphatic sounds. This means there is a difficulty in pronunciation for emphatic sounds. So, the speakers give emphatic sounds more attention compared to non-emphatic sounds.

## 3.4. A comparison between MSA Arabic with other dialects of Arabic and other languages

Before this section, we discussed comparing some stop sounds (/b/, and /k/) with some Arabic dialects and other languages. In this section, we will give more details by comparing between 14 languages and 5 Arabic dialects as we see in Table 8, where (N/A) means this sound is not found in this language and (**) means this sound does not have a computed VOT value.

Table 8. A comparison between 14 languages and 5 Arabic dialects [1][3][9][16] (milliseconds)

| Language / phonemes | /d/ | /d?/ | /t/ | /t?/ | /k/ | /b/ | /p/ | /g/ | /q/ |
|---|---|---|---|---|---|---|---|---|---|
| Portuguese | N/A | N/A | 10.8 | N/A | 15.7 | N/A | 6.7 | N/A | N/A |
| Hungarian | -87 | N/A | 16 | N/A | 29 | -90 | 2 | -58 | N/A |
| Tamil | -78 | N/A | 8 | N/A | 24 | -74 | 12 | -62 | N/A |
| Dutch | -80 | N/A | 15 | N/A | 25 | -85 | 10 | N/A | N/A |
| Spanish | -51.2 | N/A | 22.6 | N/A | 40.7 | -58.2 | 18.8 | -44 | N/A |
| Polish | -89.9 | N/A | 27.9 | N/A | 52.7 | -88.2 | 21.5 | -66.1 | N/A |
| French | N/A | N/A | 41 | N/A | 54 | N/A | 22 | N/A | N/A |
| English | 5 | N/A | 70 | N/A | 80 | 1 | 58 | 21 | N/A |
| Danish | 17 | N/A | 79 | N/A | 74 | 14 | 66 | 23 | N/A |
| Swedish | 20 | N/A | 120 | N/A | 130 | 10 | 115 | 25 | N/A |
| Korean | N/A | N/A | 11 | N/A | 19 | N/A | 7 | N/A | N/A |
| Hindi | -76 | N/A | 15 | N/A | 18 | -85 | 13 | -63 | N/A |
| Saudi dialect | ** | ** | 32 | 20 | 36 | ** | N/A | ** | N/A |
| Labanese dialect | -40 | ** | 30 | ** | 30 | -40 | N/A | N/A | 10 |
| Jordanian dialect (short vowel) | 10 | ** | 37 | ** | 39 | ** | N/A | 15 | N/A |
| Jordanian dialect (long vowel) | 23 | ** | 64 | ** | 60 | ** | N/A | 20 | N/A |
| MSA Arabic (our thesis) | 14.75 | 11.5 | 49 | 16.25 | 52 | 13 | N/A | N/A | 24.75 |
| Quran resitation (our thesis) | 17 | 10 | 36 | 12.33 | 37 | 13 | N/A | N/A | 15.67 |

We observed the voiced stop sounds in all Arabic dialects are positive VOT values except Lebanon dialect. Depending on Lisker et al. [1], we found MSA Arabic and CA Arabic in voiced sounds have short VOT while unvoiced sounds have long VOT. Also, as we know from previous researchers [19] the /d$^?$/, /t$^?$/ and /q/ sounds are found only in the Arabic language.

On the other side, we found for /d/ the maximum VOT is Quran recitation and the minimum VOT value is the Polish language and the standard deviation between them is high. For /k/ the maximum VOT is the Swedish language and the minimum VOT value is both the Korean and Hindi languages and the standard deviation is high. For /b/ we regarded the maximum VOT is MSA Arabic language and the minimum VOT value the Hungarian language and the standard deviation is high. For /p/ we regarded the maximum VOT is the Swedish language and the minimum VOT value is the Hungarian language and the standard deviation is high. For /t/ we regarded the maximum VOT is the Swedish language and the minimum VOT value is the Tamil language and the standard deviation is high. For /g/ we regarded the maximum VOT is the Swedish language and the minimum VOT value is the Polish language and the standard deviation is high.

However, the VOT value changes between Arabic dialects and languages. This result is supported by previous researchers; the VOT can be used to classify or detect for a dialect or language [1][3]. Also, supported by the previous researchers who said that the Swedish language is classified from languages, which have long VOT values and can classify the Hungarian and Polish languages as languages which have short VOT value [30][31].

### 3.5. Studying the dialect effect in MSA Arabic

In this section, we will investigate the effect of some different MSA Arabic dialects for VOT values. To study that, we selected samples of speakers as we see in Table 10. Also, we fixed the qualification as memorization of all Quran chapters and the gender as male for these speakers. Information about these speakers also is clear in the below table.

Table 10. Studying the dialect effect on MSA Arabic

| Name File | age | degree | gender | /d/ | /d?/ | /t/ | /t?/ |
|---|---|---|---|---|---|---|---|
| s01cxext1 | 17 | All | M | 17 | 16 | 55 | 16 |
| s01cxext2 |  | All |  | 19 | 9 | 63 | 15 |
| s02cxext1 | 18 | All | M | 22 | 10 | 61 | 23 |
| s02cxext2 |  | All |  | 17 | 12 | 48 | 24 |
| min |  |  |  | 17 | 9 | 48 | 15 |
| max |  |  |  | 22 | 16 | 63 | 24 |
| avg |  |  |  | 18.75 | 11.75 | 56.75 | 19.5 |
| std. dev. |  |  |  | 2.362908 | 3.095696 | 6.751543 | 4.654747 |
| s15cxext1 | 20 | All | M | 16 | 13 | 64 | 17 |
| s15cxext2 |  | All |  | 14 | 15 | 66 | 18 |
| s19cxext1 | 18 | All | M | 16 | 12 | 50 | 13 |
| s19cxext2 |  | All |  | 15 | 14 | 50 | 15 |
| min |  |  |  | 14 | 12 | 50 | 13 |
| max |  |  |  | 16 | 15 | 66 | 18 |
| avg |  |  |  | 15.25 | 13.5 | 57.5 | 15.75 |
| std. dev. |  |  |  | 0.957427 | 1.290994 | 8.698659 | 2.217356 |
| s06cxext1 | 20 | All | M | 16 | 10 | 33 | 17 |
| s06cxext2 |  | All |  | 14 | 12 | 36 | 15 |
| s21cxext1 | 19 | All | M | 16 | 12 | 55 | 20 |
| s21cxext2 |  | All |  | 19 | 12 | 52 | 19 |
| min |  |  |  | 14 | 10 | 33 | 15 |
| max |  |  |  | 19 | 12 | 55 | 20 |
| avg |  |  |  | 16.25 | 11.5 | 44 | 17.75 |
| std. dev. |  |  |  | 2.061553 | 1 | 11.10555 | 2.217356 |

From the table, these speakers are from three regions in the Kingdom of Saudi Arabia. Speakers 1 and 2 are from AlQassem region (North of the kingdom) and speakers 15 and 19 are from Jazan region (South of the kingdom) and speakers 6 and 21 are from Sader region (Center of the kingdom). We found the range of VOT value in the AlQassem region, for /d/ is between 17 to 22 milliseconds, for /d$^?$/, between 9 to 16 milliseconds, for /t/, between 48 to 63 milliseconds, and for /t$^?$/, between 15 to 24 milliseconds. Also, we found the range of the VOT value in the Jazan region, for /d/, between 14 to 16 milliseconds, for /d?/, between 12 to 15 milliseconds, for /t/ , between 50 to 66 milliseconds and for /t$^?$/, between 13 to 18 milliseconds. In addition, we found the range of the VOT value in the Sader region for /d/, between 14 to 19 milliseconds, for /d$^?$/, between 10 to 12 milliseconds, for /t/ , between 33 to 55 milliseconds and for /t$^?$/, between 15 to 20 milliseconds. We observed the VOT values of /t/ sound are more dependent on the different Arabic dialects.

After this investigation, we can say the dialect affects the VOT value. In other words, VOT value changes between dialects. This result is supported by the previous researchers, they said the VOT can be used to classified or detected for a dialect or language [1][3].

### 3.6. Boundary voiced and unvoiced stop sounds in MSA Arabic

After all these results and also depending on previous research [1][4], we know that the main goal from measuring VOT is to make a distinction between voiced stop sound and unvoiced stop sound. In this section, we will focus on the boundary for both classes in MSA Arabic language. Before this Section, we know the MSA Arabic language has positive VOT values. Also, we found the category MSA Arabic depends on Lisker et al. [1], voiced sounds have short VOT while unvoiced sounds have long VOT. To know the boundary for both classes we selected a group of speakers as we see in Table 9.

Table 9. Boundary voiced and unvoiced stop sounds in MSA Arabic

| Name File | age | degree | gander | /b/ | /d/ | /d?/ | /t/ | /t?/ | /k/ | /q/ |
|---|---|---|---|---|---|---|---|---|---|---|
| s51cxext1 | 22 | 20 | M | 11 | 17 | 11 | 52 | 22 | 34 | 18 |
| s51cxext2 | 22 | 20 |  | 11 | 15 | 11 | 52 | 22 | 34 | 17 |
| s59cxext1 | 26 | All | M | 12 | 17 | 11 | 50 | 16 | 47 | 26 |
| s59cxext2 |  | All |  | 10 | 16 | 10 | 46 | 17 | 50 | 27 |
| min |  |  |  | 10 | 15 | 10 | 46 | 16 | 34 | 17 |
| max |  |  |  | 12 | 17 | 11 | 52 | 22 | 50 | 27 |
| avg |  |  |  | 11 | 16.25 | 10.75 | 50 | 19.25 | 41.25 | 22 |
| std. dev. |  |  |  | 0.816497 | 0.957427 | 0.5 | 2.828427 | 3.201562 | 8.460693 | 5.228129 |
| s41cxext1 | 20 | 1 | F | 11 | 12 | 10 | 38 | 13 | 29 | 17 |
| s41cxext2 | 20 | 1 |  | 10 | 12 | 10 | 41 | 15 | 24 | 20 |
| s42cxext1 | 19 | 1 | F | 10 | 15 | 12 | 36 | 16 | 49 | 22 |
| s42cxext2 | 19 | 1 |  | 10 | 16 | 10 | 40 | 18 | 50 | 20 |
| min |  |  |  | 10 | 12 | 10 | 36 | 13 | 24 | 17 |
| max |  |  |  | 11 | 16 | 12 | 41 | 18 | 50 | 22 |
| avg |  |  |  | 10.25 | 13.75 | 10.5 | 38.75 | 15.5 | 38 | 19.75 |
| std. dev. |  |  |  | 0.5 | 2.061553 | 1 | 2.217356 | 2.081666 | 13.44123 | 2.061553 |
| s46cxext1 | 14 | 1 | M | 15 | 11 | 10 | 36 | 13 | 44 | 18 |
| s46cxext2 | 14 | 1 |  | 13 | 13 | 12 | 36 | 13 | 63 | 28 |
| s47cxext1 | 15 | 1 | M | 24 | 25 | 15 | 28 | 16 | 69 | 25 |
| s47cxext2 | 15 | 1 |  | 21 | 17 | 12 | 33 | 14 | 64 | 22 |
| min |  |  |  | 13 | 11 | 10 | 28 | 13 | 44 | 18 |
| max |  |  |  | 24 | 25 | 15 | 36 | 16 | 69 | 28 |
| avg |  |  |  | 18.25 | 16.5 | 12.25 | 33.25 | 14 | 60 | 23.25 |
| std. dev. |  |  |  | 5.123475 | 6.191392 | 2.061553 | 3.774917 | 1.414214 | 10.98484 | 4.272002 |

We extracted the boundary from this table as we see in Figure 5. We found the boundary average for voiced stop sounds is between 11 milliseconds and 15 milliseconds and unvoiced stop sounds are between 25 milliseconds and 36 milliseconds in MSA Arabic language.

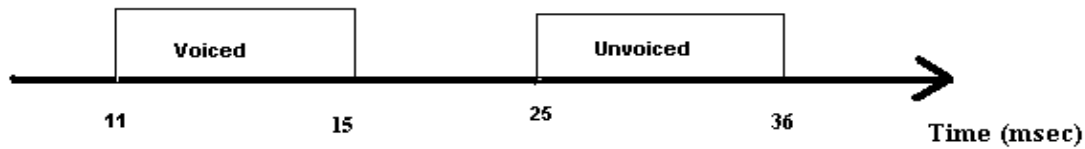

Figure 5. The boundary average VOT for voiced sounds and unvoiced stop sounds in MSA

However, we always observed the average VOT for voiced Stop sounds (/d/, /d$^?$/ and /b/) are less than the average VOT for unvoiced stop sounds (/t/, /k/ and /t$^?$/). Also, we see that the non-overlapping VOT values between voiced sounds and unvoiced sounds are 10 milliseconds. This is useful for classification of voiced and unvoiced stop sounds. In other words, from VOT value, we can determine if the stop sound is voiced or unvoiced.

Also, we know from previous research in eleven languages such as English and Spanish, the voiced sounds (negative VOT) are always less than unvoiced sounds (positive VOT). It means the vibration of vocal cords in voiced sounds start before unvoiced sounds.

## 4. CONCLUSION AND FUTURE WORK

We investigated MSA Arabic stop sounds, namely /d/, /t/, /d$^?$/ and /t$^?$/ by analyzing the VOT values, Comparing and analyzing voiced/unvoiced sounds which are /d/ and /t/, Comparing and analyzing Emphatic sounds which are /t$^?$/ and /d$^?$/, Studying the gender effect, Studying the Memorization effect, The effect of emphasis in MSA Arabic language, The effect of VOT in different MSA Arabic. In the beginning, when corpus was collected, each speaker utters seven words and the sounds were investigated. These words were chosen to confirm that the targeted sounds are in the middle of the word while the former and the later sounds with respect to the targeted sounds are always identical. The words formation is CV-CV-CV. The speaker repeats this set of words 5 times. Therefore, the total number of the recorded utterances is (60 speakers × 7 words × 5 trials = 2100 recorded words). On other side, depending on Lisker et al. [1], they divided language depending on VOT value. We know before research in eleven languages such as English, the voiced sounds (negative VOT) are always less than unvoiced sounds (positive VOT). It means the vibration of the vocal cords in voiced sounds start before unvoiced sounds. In our thesis, we ended to a conclusion that in MSA Arabic the VOT values of these stops are positive regardless of the voicing where /d/ and /d?/ are voiced sound, but /t/ and /t$^?$/ are not. Moreover, when we compared average VOT values for emphatic stop sounds (/d$^?$/ and /t$^?$/), we always found the VOT for voiced sound /t?/ was more than unvoiced sound /d$^?$/. Also, we found there was no overlapping in average VOT between them, both is 6 msec. On the other hand, we realized when there is an increase in the memorization of THQ chapters, and we are going towards a correct VOT value. Also, we always found the average value VOT in male is more than the average value VOT female. The reason is the pitch Period (P.P) – detecting the starting of vibration for vocal cord – for male take longer time compared to female. So, the VOT's of male speakers are affected by this increase of period length and due to increase in VOT values. In addition, we always found the emphatic sounds are less than non-emphatic sounds. Also, we found the standard division for non emphatic sounds is more than the standard deviation for emphatic sounds; this means the difficulty in pronunciation for emphatic sounds. So, the speakers give emphatic sounds more attention compared to non emphatic sounds. Finally, when we made comparison between MSA Arabic with other dialects of Arabic, we regarded the VOT value change between Arabic dialects and languages. This result is supported by the previous researchers, they said the VOT can be used to classify or detect for a dialect or language [1][3].

Our future prospects include further work in this field. For instance, adding more speakers, both male and female to our database. We plan to validate our results by statistical methodology. Additional investigations may include the effect of gender and noise on VOT values, a study of all the dialects in the Kingdom of Saudi Arabia to see their effects on VOT values, and possibly, a system to identify stop sounds in MSA Arabic language.

## ACKNOWLEDGMENT

Grateful acknowledgment to King Abdulaziz City for Science and Technology (KACST) for their partial support of my work.Also, acknowledgment to my supervisor Dr. Yousaf A. Alotaibi who supported me in master thesis in King Saud University (KSU). Also, acknowledgment to Eng. Saleh Alobaishi, who helped review this paper.